\pdfoutput=1

\documentclass[11pt]{article}

\usepackage[preprint]{acl}
\usepackage{times}
\usepackage{latexsym}
\usepackage{xspace}
\usepackage{amsmath}
\usepackage{amssymb}
\usepackage{booktabs}
\usepackage{multirow}
\usepackage{enumitem}
\usepackage[T1]{fontenc}
\usepackage[utf8]{inputenc}
\usepackage{microtype}
\usepackage{inconsolata}
\usepackage{graphicx}
\usepackage{makecell}

\newcommand{\name}{\textsc{LiteASR}\xspace}
\newcommand{\ie}{\textit{i.e.},\xspace}
\newcommand{\eg}{\textit{e.g.},\xspace}

\newcommand{\pgheading}[1]{\noindent\textbf{#1.}}

\newcommand{\whisperL}{Whisper large-v3\xspace}
\newcommand{\whisperT}{Whisper large-v3-turbo\xspace}
\newcommand{\whisperM}{Whisper medium\xspace}
\newcommand{\distillwhisper}{Distill-Whisper\xspace}
\newcommand{\kotobawhisper}{Kotoba-Whisper\xspace}
\newcommand{\canary}{Canary 1B\xspace}

\title{\name: Efficient Automatic Speech Recognition with\\Low-Rank Approximation}

\author{
Keisuke Kamahori\textsuperscript{1,2} \quad 
Jungo Kasai\textsuperscript{2} \quad 
Noriyuki Kojima\textsuperscript{2} \quad 
Baris Kasikci\textsuperscript{1} \\
\textsuperscript{1}University of Washington  
\textsuperscript{2}Kotoba Technologies, Inc. \\
\texttt{\{kamahori,baris\}@cs.washington.edu, \{jkasai,nkojima\}@kotoba.tech}
}

\begin{document}
\maketitle

\begin{abstract}
Modern automatic speech recognition (ASR) models, such as OpenAI’s Whisper, rely on deep encoder-decoder architectures, and their encoders are a critical bottleneck for efficient deployment due to high computational intensity.
We introduce \name, a low-rank compression scheme for ASR encoders that significantly reduces inference costs while maintaining transcription accuracy. 
Our approach leverages the strong low-rank properties observed in intermediate activations: by applying principal component analysis (PCA) with a small calibration dataset, we approximate linear transformations with a chain of low-rank matrix multiplications, and further optimize self-attention to work in reduced dimensionality. 
Evaluation results show that our method can compress \whisperL's encoder size by over 50\%, matching \whisperM's size with better transcription accuracy, thereby establishing a new Pareto frontier of accuracy and efficiency.
The code of \name is available at \url{https://github.com/efeslab/LiteASR}.
\end{abstract}

\section{Introduction}
\label{sec:intro}
Automatic speech recognition (ASR) systems have made significant strides in recent years, achieving near-human transcription performance \cite{radford2023robust,puvvada2024less}. 
Modern ASR models, such as OpenAI’s Whisper family, typically adopt an encoder-decoder architecture \cite{radford2023robust}. 
For instance, \whisperL comprises 32 Transformer blocks in both its encoder and decoder, totaling approximately 1.6 billion parameters, and has set new standards in multilingual transcription accuracy.

Despite these advances, deploying ASR systems in real-world applications poses substantial efficiency challenges. 
First, many applications, such as live transcription, voice assistants, and real-time translation, impose strict latency requirements \cite{machavcek2023turning,bevilacqua2024whispy,nguyen2020low,wang2022lamassu,jeffries2024moonshine}. 
Latency refers to the delay between the input of audio and the output of the transcribed text. 
In real-time applications, even a few seconds of delay can significantly degrade user experience. 

Second, while the overall model size may be moderate compared to the latest large language models (LLMs), ASR encoders are computationally intensive due to the long input sequences they must process. 
For instance, the encoder Transformers in the Whisper series consistently process input sequences of length 1500.
For real-time applications, this encoder must be processed frequently, making it a significant computational bottleneck.

\begin{figure}[t]
  \centering
  \includegraphics[width=\linewidth]{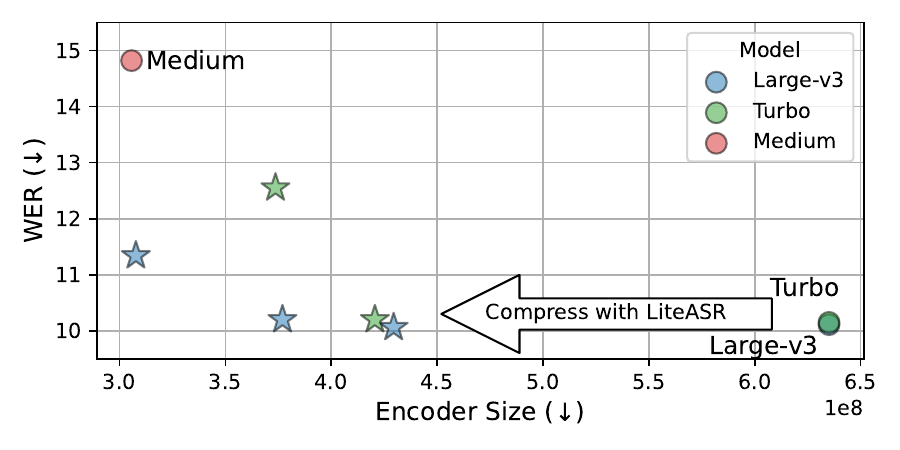}
  \caption{The relationship between encoder size and accuracy, as measured by word error rate (WER), for models in the Whisper family. The stars denote variants compressed via our method, which achieves an optimal trade-off between accuracy and efficiency.}
  \label{fig:fig1}
\end{figure}

These challenges are acute in both on-device and data center settings. 
In on-device scenarios (\eg laptops or smartphones), limited hardware capabilities make it difficult to meet latency constraints. 
Even in data center environments, which serve multiple concurrent users, the high computational intensity of ASR encoders becomes a critical bottleneck.
Although batching can improve serving throughput for memory-bound workloads, such as ASR decoders, it provides limited benefits for compute-bound encoders (as discussed in \S \ref{sec:background}).

Moreover, recent works have shown that the decoder component of ASR models can be aggressively compressed.
For example, OpenAI's \whisperT successfully reduced the number of decoder layers from 32 down to 4 layers via distillation.
Other variants, such as \distillwhisper and \kotobawhisper, have taken this even further, compressing the decoder to as few as 2 layers \cite{gandhi2023distil,kotoba-whisper-v2-0}. 
However, the encoder part remains largely unexplored, making its optimization increasingly crucial for efficient ASR systems.

In this work, we propose \name, a novel compression scheme that targets ASR encoders by exploiting the low-rank structure of hidden activations during inference. 
A key insight driving our approach is that intermediate activations, both in self-attention and multi-layer perceptron (MLP) layers, consistently exhibit low-rank properties across a wide variety of inputs. 
This phenomenon stems from ASR encoders’ use of Mel spectrograms, the 2D time-frequency audio representations. 
Real-world audio (\eg human speech) exhibits strong correlations between frequency components \cite{huang2012singing,zergat2013robust,tian2024user,kacha2020principal}, resulting in low-rank characteristics of the intermediate features.

Our method first analyzes the low-rank properties of activations using a small amount of calibration data. 
We then perform a principal component analysis (PCA) \cite{wold1987principal} to extract the dominant components and approximate linear transformations with rank-$k$ projections. 
This factorization allows each weight matrix to be expressed as the product of two lower-rank matrices, thereby reducing the total number of floating-point operations (FLOPs) required for inference. 
We employ an adaptive mechanism based on the threshold to determine the optimal degree of low-rank approximation for each layer. 

To further capitalize on the optimization, we also modify the self-attention algorithm to operate in reduced dimensionality.
We implement a specialized GPU kernel based on FlashAttention \cite{dao2022flashattention} to accelerate the computation of attention scores and outputs.

Our evaluation shows that \name achieves an optimal trade-off between accuracy and efficiency (see Figure~\ref{fig:fig1}). 
When applied to \whisperL, \name reduces the encoder size by approximately 40\%, yielding an execution speedup of around 1.4x with negligible accuracy loss. 
In alternative configurations, we further reduce the model size to less than half, resulting in a model comparable in size to \whisperM, while delivering improved accuracy.
We also demonstrate the applicability of the method across different languages and models (\S \ref{sec:experiments}).

In summary, this paper makes the following contributions:
\begin{enumerate}
    \item We introduce \name, a compression method for ASR encoders using a low-rank approximation of activation values. This method approximates linear layers with a chain of low-rank matrix multiplications and optimizes self-attention to operate in reduced dimensionality.
    \item We present a comprehensive evaluation demonstrating that our method achieves a Pareto frontier of accuracy and efficiency.
\end{enumerate}

The rest of this paper is organized as follows: \S\ref{sec:background} gives background on ASR efficiency, \S\ref{sec:methodology} presents our low-rank approximation framework, \S\ref{sec:experiments} details the experimental setup, results, and analysis, \S\ref{sec:related-work} reviews related work, and \S\ref{sec:conclusion} concludes the paper.

\section{Background}
\label{sec:background}
\subsection{Automatic Speech Recognition (ASR)}
ASR models convert spoken language into text by transforming raw audio into a compact representation, such as a Mel spectrogram, and processing it with neural networks. 
Modern systems often use encoder-decoder architectures, typically employing Transformers \cite{radford2023robust,puvvada2024less,rekesh2023fast,gulati2020conformer}. 
For instance, OpenAI's Whisper mainly uses Transformer blocks, each of which consists of self-attention and MLP layers with a large number of linear transformations (query/key/value/out projections for self-attention and two larger linear transformations for MLP).
A notable recent trend in ASR models is the reduction in decoder size without compromising performance, as exemplified by models such as \whisperT \cite{radford2023robust} and \distillwhisper \cite{gandhi2023distil}, which reduced the number of decoder layers from 32 to 4 and 2, respectively.

\begin{figure}[t]
  \centering
  \includegraphics[width=\linewidth]{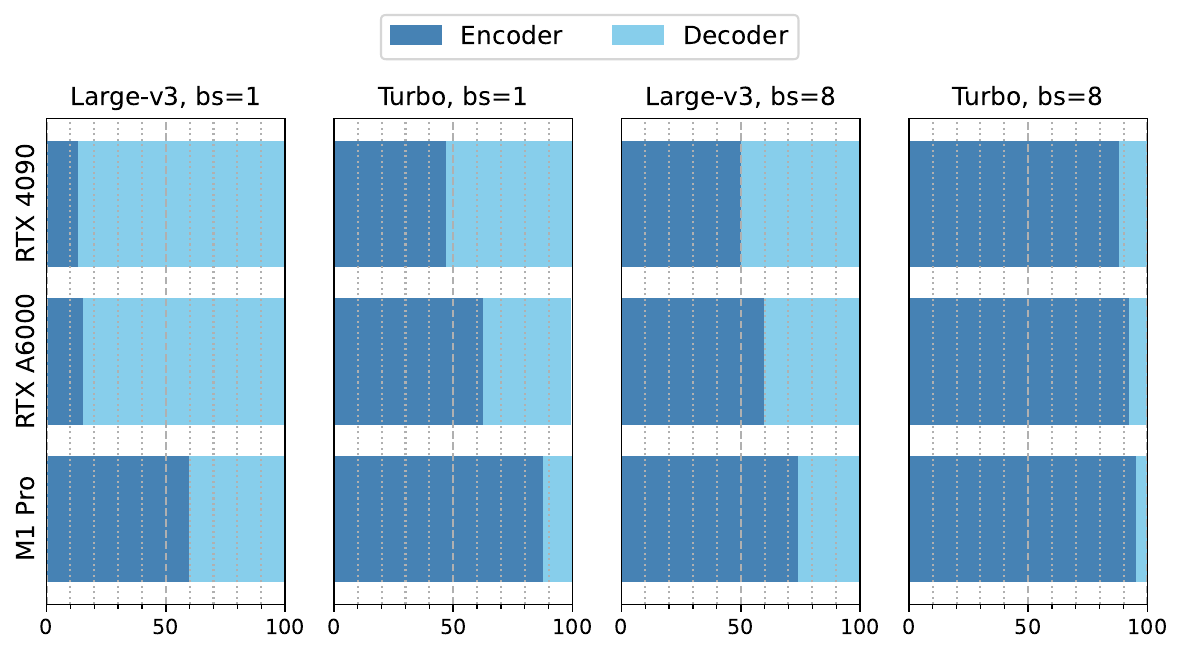}
  \caption{Latency breakdown of encoder and decoder relative to end-to-end latency for \whisperL and \whisperT models under varying batch sizes (1 and 8).}
  \label{fig:latency-breakdown}
\end{figure}

\subsection{Compute Requirements of ASR Models}
Since the encoder often processes long sequences (\eg fixed at 1500 for Whisper), it often emerges as the primary runtime bottleneck. 
Figure~\ref{fig:latency-breakdown} shows the latency breakdown between the encoder and decoder across three hardware setups (NVIDIA RTX 4090, NVIDIA RTX A6000, and Apple M1 Pro), two models (\whisperL and \whisperT), and two batch sizes (1 and 8).\footnote{We use vLLM \cite{kwon2023efficient} (ver. 0.7.0) and MLX \cite{mlx2023} (ver. 0.21.1) to transcribe a sample audio clip from the ESB dataset \cite{gandhi2022esb}.}

Although the encoder only accounts for about 15\% of the overall latency for single-batch \whisperL on GPUs, it represents a more significant bottleneck in other scenarios.
For the newer \whisperT model, the latency contribution of the encoder increases significantly due to the reduced size of the decoder. 
Similarly, for on-device inference (\eg M1 Pro), the encoder's relative latency is higher due to the limited computational power of such devices compared to GPUs.

In data center deployment scenarios where multiple requests are batched, the encoder’s latency impact is further exacerbated. 
For example, with a batch size of 8 and using \whisperT, the encoder can consume over 90\% of the total latency. 
This disproportionate latency is primarily due to the encoder's high computational intensity \cite{williams2009roofline}; batching is therefore ineffective at increasing throughput for encoders. 
In contrast, the decoder generates tokens one at a time in an autoregressive manner and is memory-bound, bottlenecked by memory bandwidth rather than computational power, making batching an effective strategy to enhance throughput \cite{lequn-blog}. 
Consequently, although batching can substantially improve serving throughput for the decoder, it offers limited benefits for the compute-bound encoder and the encoder becomes a notable bottleneck for large batch sizes.

These findings collectively highlight the encoder as a critical bottleneck for efficient ASR deployment in both on-device and data center environments. 
This issue becomes more pronounced with recent trends toward smaller decoders. 
Therefore, there is a strong demand for methods to reduce the computational requirements of the encoder.

\begin{figure}[t]
  \centering
  \includegraphics[width=\linewidth]{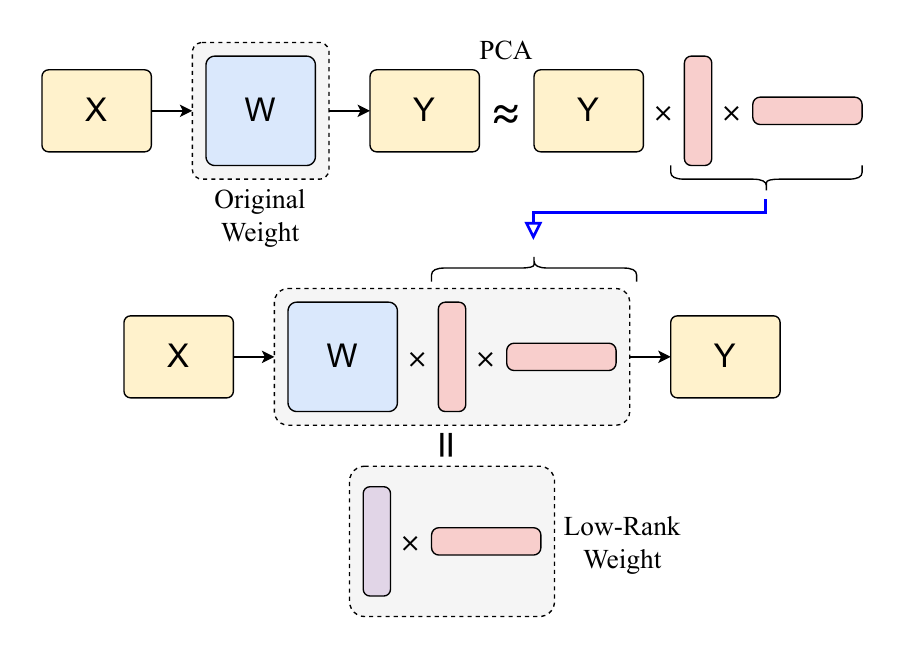}
  \caption{A simplified illustration of our proposal. We use low-rank decomposition of activation values (Y) to compress the weight (W).}
  \label{fig:method}
\end{figure}

\section{Methodology}
\label{sec:methodology}
Our method, \name, compresses the ASR encoder by extracting the low-rank features from activations at different layers of the model.
To do so, we first use calibration data to analyze activations and then convert the dense matrix multiplication within the model to the product of low-rank matrices (Figure~\ref{fig:method} shows a simplified overview of the method). 
We further modify the self-attention algorithm to work efficiently in reduced dimensionality. 
In this section, we explain the methodologies in detail.

\subsection{Analyzing Activations in Transformers}
Consider a linear layer defined by
\begin{equation}
    Y = XW + b,
\end{equation}
where the weight matrix $W \in \mathbb{R}^{D_{\text{in}} \times D_{\text{out}}}$ and the bias vector $b \in \mathbb{R}^{D_{\text{out}}}$ are learnable model parameters. 
Here, the input activations $X \in \mathbb{R}^{L \times D_{\text{in}}}$ produce the output activations $Y \in \mathbb{R}^{L \times D_{\text{out}}}$ during the forward pass. 
In this notation, $D_{\text{in}}$ and $D_{\text{out}}$ denote the input and output dimensions of the layer, respectively, and $L$ is the sequence length.\footnote{For Whisper encoders, this is always $1500$.}

To study the distribution of activations, we collect calibration data consisting of $N_{\text{calib}}$ inputs. For each linear layer, we record the corresponding output $Y$. 
The resulting dataset can be viewed as $L \times N_{\text{calib}}$ samples, where each sample is a $D_{\text{out}}$-dimensional vector. For simplicity, we refer to this collection of samples as $Y$.

Our goal is to approximate the observed activations by projecting them onto their principal components. 
First, let $Y_\mathrm{M} \in \mathbb{R}^{D_{\text{out}}}$ denote the mean vector of the dataset $Y$. Following the standard PCA procedure, we perform a singular value decomposition (SVD) on the mean-centered data:
\begin{equation}
    U, S, V = \mathrm{SVD}(Y - Y_\mathrm{M}).
\end{equation}
Here, $V \in \mathbb{R}^{D_{\text{out}} \times D_{\text{out}}}$ is the matrix of right singular vectors. By selecting the first $k$ columns of $V$, denoted by $V_k \in \mathbb{R}^{D_{\text{out}} \times k}$, we capture the top-$k$ principal components of the data. The original activations can then be approximated as:
\begin{equation}
    \label{eq:pca-approx}
    Y - Y_\mathrm{M} \approx (Y - Y_\mathrm{M})\,V_k\,V_k^\top.
\end{equation}
This approximation retains the most significant features of $Y$ while reducing its dimensionality.

\subsection{Compressing Model Layers}
Using the PCA approximation from Equation~\ref{eq:pca-approx}, we can rewrite the original linear layer $Y = XW + b$ as a combination of low-rank matrix multiplications. 
Substituting $Y = XW + b$ gives
\begin{equation}
\begin{aligned}
\label{eq:linear-decomposed}
Y - Y_\mathrm{M} &\approx (XW + b - Y_\mathrm{M})\,V_k\,V_k^\top \\
Y &\approx (XW + b - Y_\mathrm{M})\,V_k\,V_k^\top + Y_\mathrm{M}.
\end{aligned}
\end{equation}
This expression can be reorganized as
\begin{equation}
Y \approx X (W V_k) V_k^\top + \Bigl(Y_\mathrm{M} + (b - Y_\mathrm{M})\,V_k\,V_k^\top\Bigr).
\end{equation}
In this factorization, the original layer is decomposed into:
\begin{itemize}
    \item Two low-rank linear transformations, with weight matrices $W V_k \in \mathbb{R}^{D_{\text{in}} \times k}$ and $V_k^\top \in \mathbb{R}^{k \times D_{\text{out}}}$, and
    \item A constant bias term given by $Y_\mathrm{M} + (b - Y_\mathrm{M})\,V_k\,V_k^\top$.
\end{itemize}
Since both weight matrices and bias can be pre-computed using calibration data, this decomposition significantly reduces FLOPs when $k$ is much smaller than the original dimension.

\subsubsection{How to Choose $k$}
Choosing the appropriate value for $k$ involves a trade-off between accuracy and efficiency. 
A smaller $k$ leads to a more aggressive approximation, which increases efficiency but may incur a larger accuracy loss.

\paragraph{Accuracy Constraint.}
To preserve accuracy, the top-$k$ principal components must capture a sufficient portion of total variance.
Let $S \in \mathbb{R}^{D_{\text{out}}}$ denote the singular values from the SVD of the mean-centered activations (assumed to be sorted in decreasing order). We enforce
\begin{equation}
\label{eq:constr-acc}
\sum_{i=1}^k S_i^2 \;>\; \theta \sum_{i=1}^{D_{\text{out}}} S_i^2,
\end{equation}
where $\theta$ is a threshold that controls the trade-off between accuracy and efficiency (\ie the extent of data compression).

\paragraph{Efficiency Constraint.}
The original linear layer requires $\mathcal{O}(L D_{\text{in}} D_{\text{out}})$ FLOPs for its matrix multiplication. 
In contrast, the decomposed form in Equation~\ref{eq:linear-decomposed} requires $\mathcal{O}(L D_{\text{in}} k + L k D_{\text{out}})$ FLOPs. 
To ensure that our approximation results in a reduction of computation, we require
\begin{equation}
\label{eq:constr-flops}
L D_{\text{in}} k + L k D_{\text{out}} < L D_{\text{in}} D_{\text{out}},
\end{equation}
which simplifies to
\begin{equation}
k(D_{\text{in}} + D_{\text{out}}) < D_{\text{in}}D_{\text{out}}.
\end{equation}

For example, in \whisperL, the dimensions for self-attention layers are $(D_{\text{in}}, D_{\text{out}}) = (1280, 1280)$, and for MLP layers they are $(1280, 5120)$ or $(5120, 1280)$. 
This implies that the efficiency constraint requires $k < 640$ for self-attention and $k < 1024$ for MLP layers.

\paragraph{Practical Considerations.}
To maximize the GPU efficiency, we further restrict $k$ to be a multiple of $16$. 
Therefore, we choose $k$ as the smallest multiple of $16$ that satisfies both Equation~\ref{eq:constr-acc} and \ref{eq:constr-flops}. 
We empirically find that $\theta$ values between $0.99$ and $0.999$ achieve a good balance between accuracy and efficiency.
A detailed sensitivity study on the choice of $\theta$ is provided in \S \ref{sec:experiments}.

\subsubsection{Optimizing Self-Attention}
Moreover, there is a potential to optimize the self-attention layers further. 
Specifically, if the rank $k$ is smaller than the per-head dimension, we can compute the attention score and the value projection in alternative ways to reduce the FLOPs requirement while preserving the mathematical operations.

\paragraph{Standard Self-Attention.}
For multi-head attention, let $D_{\text{head}}$ denote the dimension per head and $h$ the number of heads (\ie the total model dimension is $D_{\text{head}} \times h$). In the $i$-th head, given an input activation matrix $X \in \mathbb{R}^{L \times D_{\text{in}}}$, the self-attention mechanism first computes three linear projections:
\begin{equation}
Q_i = X W_Q^i,\quad K_i = X W_K^i,\quad V_i = X W_V^i,
\end{equation}
where $W_Q^i,\,W_K^i,\,W_V^i \in \mathbb{R}^{D_{\text{in}} \times D_{\text{head}}}$ are the corresponding weight matrices. The standard attention output is then given by
\begin{equation}
\label{eq:attn-original}
\text{Attention}(Q_i, K_i, V_i) = \text{softmax}\!\left(\frac{Q_iK_i^\top}{\sqrt{D_{\text{head}}}}\right) V_i,
\end{equation}
with the softmax applied row-wise.
 
\pgheading{Attention Score Computation}
Using our low-rank approximation, we can factorize each projection as follows:
\begin{equation}
\begin{aligned}
Q_i &= (XW_{Q_1})W_{Q_2}^i + b_Q^i, \\
K_i &= (XW_{K_1})W_{K_2}^i + b_K^i, \\
V_i &= (XW_{V_1})W_{V_2}^i + b_V^i,
\end{aligned}
\end{equation}
where $W_{Q_1}\in \mathbb{R}^{D_{\text{in}} \times k_Q}$, $W_{Q_2}^i\in \mathbb{R}^{k_Q \times D_{\text{head}}}$, and $b_Q^i \in \mathbb{R}^{D_{\text{head}}}$ are parameters relevant for $i$-th head after low-rank approximation (with analogous definitions for $K$ and $V$). 
Here, $k_Q$, $k_K$, and $k_V$ are the respective rank sizes. 
For brevity, let $A = XW_{Q_1}$ and $B = XW_{K_1}$.
Expanding the product $Q_iK_i^\top$, we obtain:
\begin{equation}
\begin{aligned}
\label{eq:attn-score-factorized}
Q_iK_i^\top &= \Bigl( A W_{Q_2}^i + b_Q^i \Bigr)
\Bigl( B W_{K_2}^i + b_K^i \Bigr)^\top \\
&= \Bigl( A W_{Q_2}^i + b_Q^i \Bigr)
\Bigl( W_{K_2}^{i\,\top} B^\top + b_K^{i\,\top} \Bigr) \\
&= A W_{Q_2}^i W_{K_2}^{i\,\top} B^\top \\
&\quad + A W_{Q_2}^i b_K^{i\,\top} + b_Q^i W_{K_2}^{i\,\top} B^\top + b_Q^i b_K^{i\,\top}.
\end{aligned}
\end{equation}
In this expansion, the term $A W_{Q_2}^i W_{K_2}^{i\,\top} B^\top$ dominates the computational cost, while the other three terms are bias contributions.

The standard approach (Equation~\ref{eq:attn-original}) computes $Q_i$ and $K_i^\top$ separately and then multiplies them, which requires approximately $\mathcal{O}(L^2 D_{\text{head}})$ FLOPs. 
In contrast, Equation~\ref{eq:attn-score-factorized} allows us to first compute the smaller matrix product $W_{Q_2}^i W_{K_2}^{i\,\top}$ and then multiply by $A$ and $B$, reducing the computational cost to $\mathcal{O}(L\, k_Q\, k_K + L^2\, \min(k_Q, k_K))$. 
This is beneficial when $\min(k_Q, k_K) < D_{\text{head}}$.\footnote{We take the minimum of $k_Q$ and $k_K$ because we can choose the multiplication order to minimize computation.}
Thus, we adopt Equation~\ref{eq:attn-score-factorized} if the rank is sufficiently small.

\pgheading{Value projection}
After computing the attention score matrix 
\begin{equation}
S_i = \text{softmax}\!\left(\frac{Q_iK_i^\top}{\sqrt{D_{\text{head}}}}\right) \in \mathbb{R}^{L \times L},
\end{equation}
the final output is obtained by multiplying $S_i$ with $V_i$:
\begin{equation}
\begin{aligned}
S_iV_i &= S_i \Bigl((XW_{V_1}) W_{V_2}^i + b_V^i\Bigr) \\
&= S_i (XW_{V_1}) W_{V_2}^i + S_i b_V^i.
\end{aligned}
\end{equation}
Conventionally, one would first compute $(XW_{V_1}) W_{V_2}^i$ and then multiply by $S_i$, which would cost $\mathcal{O}(L^2 D_{\text{head}} + L\, k_V\, D_{\text{head}})$ FLOPs. 
However, by computing $S_i (XW_{V_1})$ first, the cost becomes $\mathcal{O}(L^2 k_V + L\, k_V\, D_{\text{head}})$ FLOPs, making this approach more efficient when $k_V < D_{\text{head}}$.

Moreover, since each row of $S_i$ sums to $1$, the bias term simplifies:\footnote{Here, $b_V^i$ is broadcast across each row of $S_i$.}
\begin{equation}
S_i b_V^i = b_V^i.
\end{equation}
Thus, the value projection can be rewritten as:
\begin{equation}
\label{eq:optimized-v-proj}
S_iV_i = \Bigl(S_i (XW_{V_1})\Bigr) W_{V_2}^i + b_V^i,
\end{equation}
which is more efficient if $k_V < D_{\text{head}}$.

\pgheading{Implementation}
To efficiently execute the operations in Equations~\ref{eq:attn-score-factorized} and \ref{eq:optimized-v-proj}, we implement a specialized kernel using Triton~\cite{tillet2019triton}. This kernel extends the original FlashAttention implementation~\cite{dao2022flashattention} to handle our optimized computation strategy.

\begin{table*}[t]
\centering
\resizebox{\textwidth}{!}{
  \begin{tabular}{cc*{9}{c}cc}
    \toprule
    \multirow{2}{*}{Model} & \multirow{2}{*}{Config.} & \multicolumn{9}{c}{WER (↓)} & \multirow{2}{*}{Size (↓)} \\
    \cmidrule(lr){3-11} & & VP & AMI & E22 & GS & LS-C & LS-O & SG & TED & Avg. & \\
    \midrule
    \multirow{4}{*}{Large-v3}
             & Original          & 8.8  & 25.9 & 19.5 & 11.1 & 2.4 & 5.5  & 3.3 & 4.4 & 10.1 & 635M (100.0\%) \\
             & \textbf{\name (a)} & 8.7  & 25.7 & 18.9 & 11.1 & 2.5 & 5.0  & 3.4 & 5.1 & \textbf{10.1} & 429M (67.6\%) \\
             & \textbf{\name (b)} & 8.4  & 28.7 & 15.8 & 12.0 & 2.7 & 6.1  & 3.1 & 4.8 & 10.2 & 377M (59.4\%) \\
             & \textbf{\name (c)} & 8.7  & 33.4 & 17.2 & 12.3 & 2.8 & 7.4  & 3.5 & 5.4 & 11.3 & \textbf{308M (48.5\%)} \\
    \midrule
    \multirow{4}{*}{Turbo}
             & Original          & 9.5  & 26.8 & 17.4 & 11.4 & 2.6 & 5.5  & 3.8 & 4.3 & 10.1 & 635M (100.0\%) \\
             & \textbf{\name (a)} & 9.0  & 27.7 & 17.0 & 11.4 & 2.8 & 6.2  & 3.1 & 4.5 & \textbf{10.2} & 421M (66.2\%) \\
             & \textbf{\name (b)} & 8.9  & 43.2 & 16.7 & 11.7 & 3.1 & 7.8  & 4.0 & 5.0 & 12.6 & 374M (58.8\%) \\
             & \textbf{\name (c)} & 10.8 & 69.7 & 35.1 & 16.0 & 4.2 & 13.7 & 5.0 & 6.4 & 20.1 & \textbf{313M (49.3\%)} \\
    \midrule
    Medium   & Original          & 8.7  & 31.3 & 25.9 & 25.9 & 3.9 & 8.8  & 5.9 & 8.2 & 14.8 & 306M (48.1\%) \\
    \bottomrule
  \end{tabular}
}
\caption{Accuracy measured by WER percentages on ESB benchmarks and encoder sizes across different configurations. Abbreviations: VP (VoxPopuli), AMI (AMI), E22 (Earnings-22), GS (GigaSpeech), LS-C (LibriSpeech test.clean), LS-O (LibriSpeech test.other), SG (SPGISpeech), TED (TED-LIUM). For encoder size, we show relative size against the original \whisperL inside parenthesis.}
\label{tab:results-acc}
\end{table*}

\section{Experiments}
\label{sec:experiments}
In this section, we describe our experimental setup and results, focusing on both the accuracy and efficiency of \name. 

\subsection{Setup}
Our primary accuracy evaluation focuses on compressing \whisperL and \whisperT, both of which have encoders of the same size. 
We use test data from the End-to-end Speech Benchmark (ESB) \cite{gandhi2022esb}, a comprehensive collection of English ASR benchmarking datasets, to assess the word error rate (WER) of both the compressed and original models. 
We randomly choose 1000 audio clips from each of the eight subsets of ESB: VoxPopuli, AMI, Earnings-22, GigaSpeech, LibriSpeech (test.clean and test.other), SPGISpeech, and TED-LIUM. 
For the calibration data, we randomly select 100 clips (non-overlapping with the test data), and the calibration process is completed within 10 minutes using a single RTX 4090 GPU. 
We employ greedy sampling with a temperature set to 0.

We present three configurations of $\theta$ for different deployment requirements: (a) \textbf{Quality-Focused}: $\theta = 0.999$ for all layers. (b) \textbf{Balanced}: $\theta = 0.99$ for self-attention layers and $\theta = 0.999$ for MLP layers. (c) \textbf{Efficiency-Focused}: $\theta = 0.99$ for self-attention layers and $\theta = 0.995$ for MLP layers.
Later, we conduct a sensitivity study for different values of $\theta$, languages, and models.

Regarding efficiency, we evaluate the encoder latency on NVIDIA RTX 4090, NVIDIA RTX A6000, and Apple M1 Pro. 
For GPUs, we modify OpenAI's Whisper implementation\footnote{\url{https://github.com/openai/whisper}} to use CUDA Graph with PyTorch \cite{ansel2024pytorch} (ver. 2.5.1), and we use Triton \cite{tillet2019triton} (ver. 3.2.0) for a customized self-attention GPU kernel. 
On the Apple device, we use MLX \cite{mlx2023} (ver. 0.21.1). 
The presented latency data are averaged over 10 runs.
Note that the encoder always takes fixed-length audio as input, so the computational requirements are exactly the same for different data.

\subsection{Accuracy Evaluation}
Table~\ref{tab:results-acc} compares the WER and encoder size. 
\name is evaluated on \whisperL and \whisperT models, with \whisperM as a reference. 
The quality-focused configuration (a) cuts model size by over 30\% with an increase of less than 0.1 percentage points in WER for both \whisperL and \whisperT.
For more efficiency-focused scenarios, configuration (b) reduces encoder size by over 40\% with comparable WER for \whisperL, and about 2.5 points degradation for \whisperT. 
Configuration (c) compresses \whisperL model to less than half, matching \whisperM's size, with better WER by about 3.5 points. 
Overall, \name significantly reduces the model size while largely maintaining accuracy.
We emphasize that, unlike typical knowledge distillation methods which require substantial data and compute,\footnote{For example, \distillwhisper~\cite{gandhi2023distil} used >20k audio hours and the distillation took days on TPU v4-8.} our approach achieves significant compression without accuracy degradation using only 100 audio clips and under 10 minutes on a single GPU.

\begin{figure}[t]
  \centering
  \includegraphics[width=\linewidth]{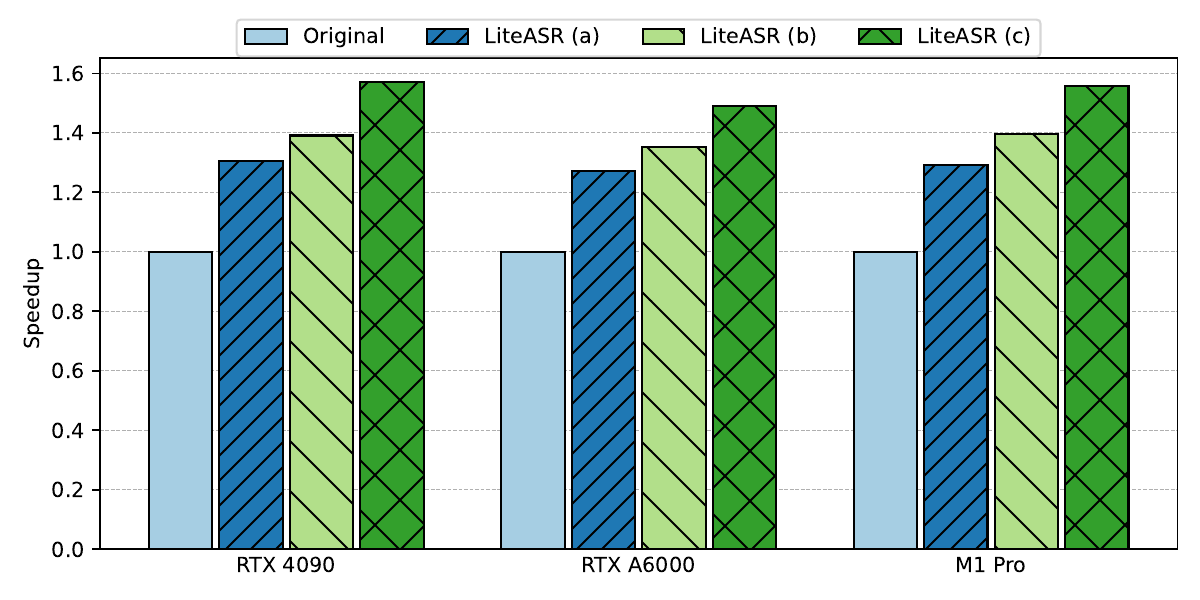}
  \caption{Execution speed of the encoder in \whisperL, compared as a ratio to the original model.}
  \label{fig:speedup}
\end{figure}

\subsection{Efficiency Evaluation}
Figure~\ref{fig:speedup} presents the efficiency evaluation results, measuring the speedup of end-to-end latency of the encoder execution compared to the original model. 
\name consistently achieves latency improvements across all three hardware setups, with average speedups of 1.29x for (a), 1.38x for (b), and 1.54x for (c). 
The best performance is observed with the RTX 4090 using (c), reaching a peak speedup of 1.57x.

\begin{figure}[t]
  \centering
  \includegraphics[width=\linewidth]{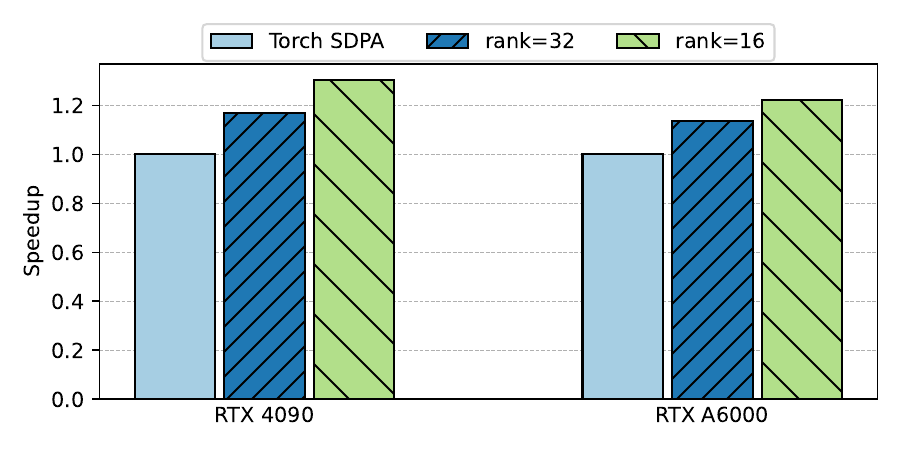}
  \caption{Our Triton kernel's performance against PyTorch implementation.}
  \label{fig:kernel}
\end{figure}
Moreover, Figure~\ref{fig:kernel} compares our Triton kernel's performance with PyTorch's scaled dot product attention (SDPA) implementation in \whisperL's encoder self-attention layers. 
The RTX 4090 shows roughly 17\% and 30\% improvements over baseline, while the RTX A6000 exhibits gains of approximately 14\% and 22\% for matrix ranks 32 and 16, respectively (\ie $k_Q$, $k_K$, $k_V$ in \S \ref{sec:methodology}, assuming all share the same value).

\begin{figure}[t]
  \centering
  \includegraphics[width=\linewidth]{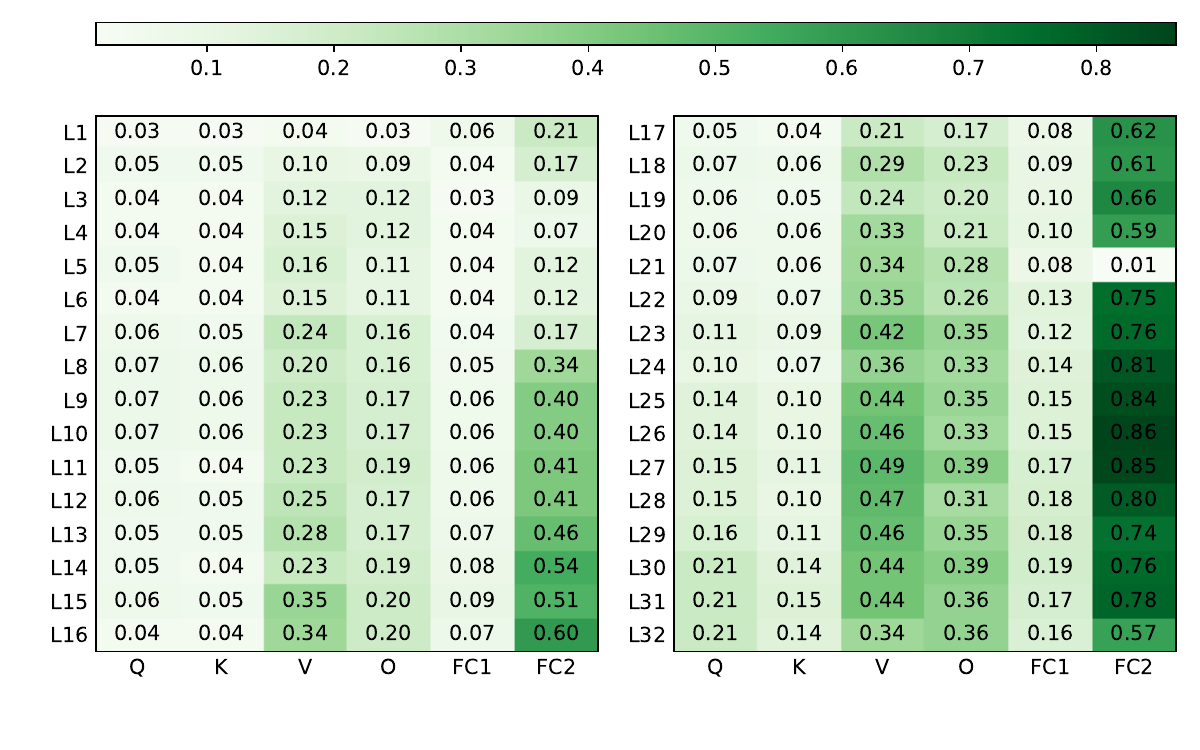}
  \caption{Compression ratio for each linear layer of \whisperL. Smaller values mean more aggressive compression}
  \label{fig:comp-ratio}
\end{figure}

\subsection{Analysis}
\subsubsection{Compression Ratio per Layer}
Figure~\ref{fig:comp-ratio} illustrates the compression ratio (\ie defined as the quotient of $k$ divided by the original dimension size) for each linear layer within the \whisperL encoder. 
The data are presented for configuration (c). 
In general, the initial layers allow for more substantial compression, with some exceptions, such as the FC2 stage in layer 21. 
This tendency is most pronounced in FC2 layers, where the earlier layers exhibit a compression ratio of less than 0.2, whereas the subsequent layers reach values larger than 0.8. 
Among the layers, the Q/K projection and FC1 layers display a smaller compression ratio compared to other layers.

\begin{figure}[t]
  \centering
  \includegraphics[width=\linewidth]{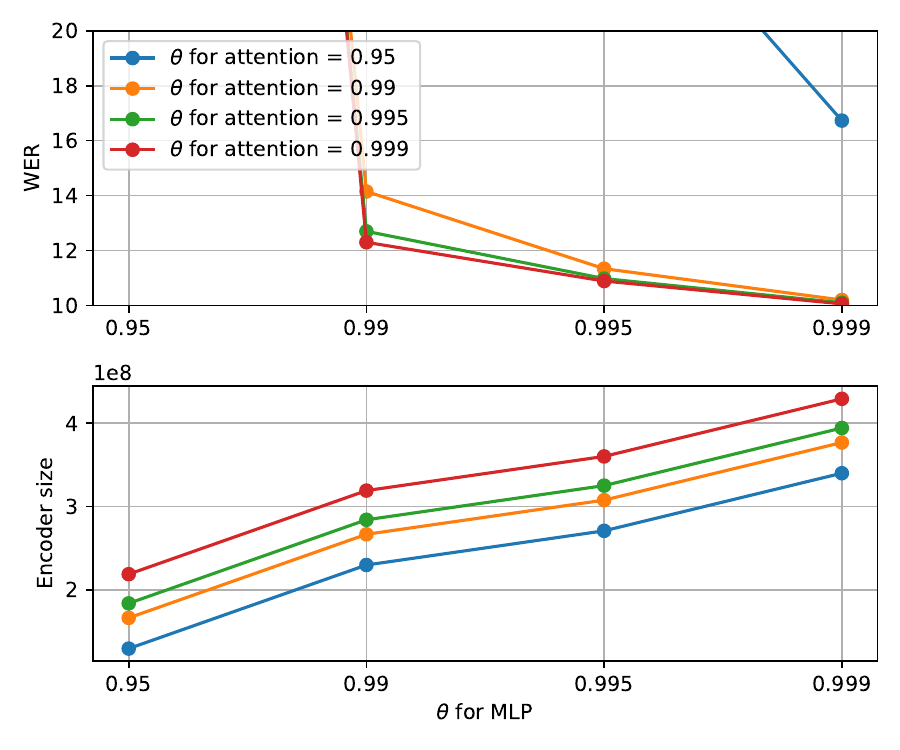}
  \caption{Sensitivity of WER and encoder size on the value of $\theta$.}
  \label{tab:sensitivity-theta}
\end{figure}

\subsubsection{Sensitivity to $\theta$}
Figure~\ref{tab:sensitivity-theta} analyzes the sensitivity of the average WER and encoder size to $\theta$ by independently varying $\theta$ from 0.95 to 0.999 for self-attention and MLP layers in \whisperL. 
Our results show a significant increase in WER when $\theta$ is below 0.99 for both layers. 
In contrast, WER improves as $\theta$ increases, with $\theta = 0.999$ achieving the best performance. 
The encoder size exhibits the opposite trend, positively correlated with $\theta$ in a steady and roughly linear fashion.
In the extreme scenario where $\theta = 0.95$ is applied to both layers, the encoder size can be reduced by around 80\%, although this comes with a significant increase in WER.

\begin{table}[t]
\centering
\begin{tabular}{c*{3}{c}c}
    \toprule
    \multirow{2}{*}{Config} & \multicolumn{2}{c}{WER (↓)} & CER (↓) & \multirow{2}{*}{Size (↓)} \\
    \cmidrule(lr){2-4} & FR & DE & JA & \\
    \midrule
     Original         & 7.2 & 13.2 & 10.8 & 635M \\
    \textbf{\name (a)} & 7.4 & 8.7 & 10.7 & 429M \\
    \textbf{\name (b)} & 6.8 & 7.7 & 11.2 & 377M \\
    \textbf{\name (c)} & 9.1 & 10.1 & 12.4 & 308M \\
    \bottomrule
\end{tabular}
\caption{Sensitivity study on other languages. Abbreviations: FR (French), DE (German), JA (Japanese).}
\label{tab:sensitivity-lang}
\end{table}

\subsubsection{Sensitivity to Languages}
To further investigate how \name generalizes to out-of-distribution data and its sensitivity to languages, we extend our evaluation to non-English benchmarks. 
We use MLS \cite{pratap2020mls} for French and German, and the JSUT basic5000 \cite{sonobe2017jsut} for Japanese.\footnote{For Japanese, we use the character error rate (CER) instead of the WER since Japanese does not have explicit word boundaries.}
Here, we use the same English calibration data as in previous experiments to compress \whisperL, and evaluate its accuracy on non-English audio. 
The results presented in Table~\ref{tab:sensitivity-lang} demonstrate \name's robustness: for (a), there is almost no degradation in accuracy, and even for (c), the degradation is less than 2 percentage points in WER/CER. 
In some cases, such as with German, we even observe an improvement in accuracy.

\begin{table}[t]
\centering
\begin{tabular}{ccc}
    \toprule
    Config & WER (↓) & Size (↓) \\
    \midrule
    Original          & 9.1 & 609M (100.0\%) \\
    \textbf{\name (a)} & 9.1 & 593M (97.3\%) \\
    \textbf{\name (b)} & 9.1 & 579M (95.0\%) \\
    \textbf{\name (c)} & 9.1 & 545M (89.4\%)  \\
    \bottomrule
\end{tabular}
\caption{Accuracy and encoder size with \canary model.}
\label{tab:sensitivity-model}
\end{table}

\subsubsection{Sensitivity to Models} 
We also evaluate on \canary \cite{puvvada2024less}, NVIDIA's state-of-the-art ASR model, to determine \name's applicability to a broader range of models. 
The encoder of Canary employs the FastConformer architecture \cite{rekesh2023fast}, and our optimizations are confined to linear layers within the feed-forward and self-attention modules, leaving the convolution modules unaltered. 
Table~\ref{tab:sensitivity-model} presents the encoder size and the average WER for ESB datasets. 
The data indicate that there is minimal degradation in the WER, although the reduction in size is moderate compared to the Whisper models, achieving approximately a 10\% reduction for configuration (c).

\section{Related Work}
\label{sec:related-work}
\subsection{Efficient ASR Inference}
Several prior works have aimed to enhance the efficiency of ASR models. 
FasterWhisper uses optimized inference kernels \cite{faster-whisper}, while WhisperX further improves it for long-form audio \cite{bain2023whisperx}. 
Whisper.cpp is a C/C++ implementation for portability on both the CPU and GPU \cite{whisper.cpp}. 
Whisper\_streaming supports live transcription for streaming purposes \cite{machavcek2023turning}. 
NVIDIA's NeMo is a modular toolkit for deploying speech models \cite{NeMo}. 
However, they do not effectively reduce ASR encoder computational demands. 
Some works provide model weight quantization, but they are limited to weights (weight-only quantization) and do not accelerate the compute-bound encoder inference.
Our approach can be integrated with these frameworks. 

Various studies, including \whisperT, \distillwhisper, and \kotobawhisper use distillation techniques to shrink decoder size \cite{radford2023robust,gandhi2023distil,kotoba-whisper-v2-0}. 
Other approaches combine distillation with quantization or lightweight modular ASR fine-tuning for underrepresented languages \cite{shao2023whisper,ferraz2023distilwhisper}.
Our work complements these efforts by further reducing the encoder's computational requirements.
Unlike these methods that require substantial data and compute resources, often tailored for specific downstream tasks, our work incurs minimal data/compute overhead. Our work also complements them by further reducing the encoder's computational requirements; for instance, our compression technique can be effectively applied to models like \whisperT, which is derived from the larger \whisperL model.

\subsection{Model Compression with Low-Rank Approximation}
The low-rank approximation has been used to compress machine learning models, such as for parameter-efficient fine-tuning \cite{hu2021lora} or the LLM's KV cache compression \cite{liu2024deepseek,chang2024palu}.
\citet{yu2023compressing} have suggested that activations in Transformer models exhibit low-rank and compressed models, mainly targeting vision models.
However, their method is limited to linear layers, leaving self-attention layers unoptimized, and its applicability to speech models has not been studied.
Another work by \citet{winata2020lightweight} presented a model architecture that adopts low-rank weights at train-time, whereas we propose a post-training technique for compressing pre-trained ASR models by analyzing activation patterns, avoiding costly retraining. 

\section{Conclusion}
\label{sec:conclusion}
In this work, we introduced a compression method for ASR encoders that leverages the inherent low-rank structure of activations in linear layers. 
By applying the PCA algorithm, this method approximates linear layers with a chain of low-rank matrix multiplications and optimizes self-attention to operate in reduced dimensionality.
Our comprehensive evaluation demonstrates that our method achieves a Pareto frontier of accuracy and efficiency, paving the way for more efficient ASR deployments for both on-device and data center environments.

\section{Limitations} 
Our method focuses on compressing linear layers and the self-attention mechanism, yielding substantial improvements for Whisper models. 
However, other architectures, such as the Conformer, include additional components such as convolution layers, which may provide further compression opportunities (see \S \ref{sec:experiments}). 
Additionally, our evaluation is currently limited to standard benchmarks in English and a few other major languages; evaluating performance on low-resource languages and domain-specific applications remains an important direction for future research. 
Finally, while our improvements do not introduce new risks per se, the enhanced efficiency could accelerate the broader adoption of ASR systems, which may amplify concerns related to privacy, surveillance, or inherent biases in large-scale deployments.

\section{Ethics Statement}
All data and models used in this paper are publicly accessible and are distributed under Creative Commons, Apache-2.0, MIT, or other open-source licenses that permit research use.

\bibliography{main}

\appendix

\section{Additional Sensitivity Studies}
\label{sec:appendix}

\subsection{Comparison to Other Baselines}
\label{sec:appndx:baseline}
Unlike \name, which compresses the model with a minimal amount of data and compute during post-training, most of the existing works on pruning or knowledge distillation for speech models require a substantial amount of data and compute, often tailored for the downstream tasks \cite{gandhi2023distil,shao2023whisper,kotoba-whisper-v2-0,someki2025context}. Therefore, they are not directly comparable with our method. 
Rather, they are complementary to us; for example, \whisperT is trained based on a larger \whisperL model, and we can apply our method on top of it.

One compression approach that works similarly to \name regarding data/compute requirements is quantization, which can synergize with our approach. Table \ref{tab:sensitivity-quant} presents the performance following int8 quantization of FasterWhisper~\cite{faster-whisper}, applied on top of the original and compressed versions of \whisperL. The results demonstrate a slight reduction in accuracy for both the original and our compressed models due to quantization, with a marginally more pronounced effect on the compressed version. Nevertheless, our compressed model maintains a high accuracy level, outperforming alternatives such as the \whisperM.

It is important to note that the quantization implementation in FasterWhisper targets only the model weights, leaving activation values at high precision. This strategy effectively reduces memory footprint but does not inherently accelerate compute-bound encoders, where performance is primarily dictated by arithmetic operation throughput rather than memory bandwidth. Our method, therefore, offers distinct advantages and can be combined with such quantization schemes for comprehensive efficiency gains.

\begin{table}[t]
\centering
\begin{tabular}{cccc}
    \toprule
    Config & Avg. WER (↓) & Size (↓) \\
    \midrule
    Original, FP16 & 10.1 & 635M \\
    Original, INT8 & 10.2 & 635M \\
    \textbf{LiteASR (b)}, FP16 & 10.2 & 377M \\ 
    \textbf{LiteASR (b)}, INT8 & 11.0 & 377M \\ 
    \bottomrule
\end{tabular}
\caption{Accuracy with different weight quantization configurations.}
\label{tab:sensitivity-quant}
\end{table}

\begin{table}[t]
\centering
\begin{tabular}{cccc}
    \toprule
    Domain & Quantity & Avg. WER (↓) & Size (↓) \\
    \midrule
    EN       &  10 & 11.4 & 351M \\
    EN       & 100 & 10.2 & 377M \\
    EN       & 200 & 10.2 & 382M \\
    \midrule 
    EN-Clean & 100 & 11.0 & 371M \\
    EN-Noisy & 100 & 10.8 & 360M \\
    FR       & 100 & 10.8 & 374M \\ 
    DE       & 100 & 13.1 & 378M \\ 
    JA       & 100 & 12.8 & 344M \\
    \bottomrule
\end{tabular}
\caption{Accuracy with different choices on calibration data quantity and domain (language).}
\label{tab:sensitivity-calib-data}
\end{table}

\subsection{Sensitivity to Calibration Data Selection}
\label{sec:appndx:calib-data}
In our experiments in \S \ref{sec:experiments}, we randomly select 100 audio clips from the English-only ESB dataset to serve as calibration data. Here, we show a sensitivity study on different aspects of calibration data selection. For these experiments, we employ \whisperL with the balanced setting (configuration (b)), varying the quantity and the selection method of the calibration data, reported in Table \ref{tab:sensitivity-calib-data}.

\textbf{Quantity.} First, we vary the number of calibration audio samples by randomly selecting 10, 100, and 200 samples from the ESB dataset, same as the main experiments (denoted as EN in the Table \ref{tab:sensitivity-calib-data}). The results indicate that using only 10 samples is insufficient, as they show a worse WER than 100 samples by more than 1 point. However, beyond 100 samples, increasing the amount is of little benefit, with the average WER remaining almost the same between 100 and 200 samples and the encoder size differing by only about 1\%.

\textbf{Domain.} Next, to evaluate the effect of data domain, we examine the impact of audio quality by selecting calibration data from different subsets of the ESB dataset, each exhibiting distinct levels of audio noise. Rather than sampling uniformly across the entire ESB dataset, we focus on the LibriSpeech test.clean subset and the AMI subset, which consistently show the lowest and highest WER, respectively, representing clean and noisy audio sources (denoted as EN-Clean and EN-Noisy in Table \ref{tab:sensitivity-calib-data}). We also compare results using calibration data sampled from non-English sources, for which we select 100 audio samples from the MLS (French, German) or JSUT (Japanese) dataset.

While the calibration data from the noisy subset yields slightly better performance than that from the clean subset, the original mixed setting (combining clean and noisy data) achieves significantly higher accuracy. This result shows the importance of utilizing diverse calibration datasets during model compression to preserve the original model's performance. In terms of language, although French outperforms both German and Japanese, non-English calibration data generally underperforms compared to the original ESB dataset. This is likely because the original model was trained with English as the primary language.

\textbf{Randomness.} The results presented in Table \ref{tab:results-acc} are from a single instance. To assess the robustness against randomness in calibration data selection, we conduct additional experiments in which the compression is run five times with different random seeds. For \whisperL with configuration (b), the mean WER across five runs is 10.15\% with a standard deviation of 0.16\%, while the average number of parameters is 378.5M with a standard deviation of 0.7M. These low standard deviations demonstrate that \name is robust to calibration dataset randomness, with minimal impact on both WER and compression ratio.

\begin{table}[t]
\centering
\begin{tabular}{ccccc}
    \toprule
    $\sigma$ & SNR & \makecell[c]{WER\\Original} & \makecell[c]{WER\\\textbf{Ours (a)}} & \makecell[c]{WER\\\textbf{Ours (b)}} \\
    \midrule
    0.01 & 14.7 & 3.1  & 3.3  & 3.4  \\
    0.02 & 8.7  & 4.2  & 4.7  & 5.3  \\
    0.03 & 5.1  & 6.6  & 6.9  & 8.5  \\
    0.04 & 2.7  & 9.6  & 10.6 & 13.0 \\
    0.05 & 0.7  & 13.2 & 14.8 & 18.3 \\
    0.06 & -0.9 & 18.2 & 20.2 & 24.9 \\
    0.07 & -2.2 & 23.2 & 26.1 & 33.3 \\
    0.08 & -3.4 & 29.0 & 33.5 & 40.4 \\
    0.09 & -4.4 & 35.7 & 40.2 & 48.6 \\
    0.1  & -5.3 & 46.7 & 47.4 & 56.2 \\
    \bottomrule
\end{tabular}
\caption{Transcription accuracy with different degrees of noise.}
\label{tab:sensitivity-snr}
\end{table}

\subsection{Robustness of Compressed Models for Noisy Data}
\label{sec:appndx:robustness}
To evaluate the robustness of \name against noisy audio data, we start from the clean data of the LibriSpeech test.clean subset, and add noise to evaluate the relation between WER and Signal-to-Noise Ratio (SNR). We inject zero-mean Gaussian noise into normalized audio by specifying a noise standard deviation ($\sigma$), producing samples that range from nearly clean to heavily corrupted, and compute the resulting SNR in decibels by comparing the average power of the clean waveform to that of the added noise. We select $\sigma$ between 0.01 and 0.1. A larger SNR value means cleaner audio. For the evaluation, we use the original and compressed versions (configurations (a) and (b)) of \whisperL.

Table \ref{tab:sensitivity-snr} presents the results. For both the original and compressed models, transcription accuracy degrades for larger noise level $\sigma$ and smaller SNR values. While the compressed model maintains comparable performance under mild noise conditions, it demonstrates reduced robustness to severe noise compared to the original model. Additionally, there is a trade-off between noise robustness and inference efficiency: conservative compression preserves greater noise tolerance, while aggressive compression prioritizes efficiency at the cost of robustness.

\begin{table}[t]
\centering
\begin{tabular}{ccc}
    \toprule
    \makecell[c]{Config \\ ($\theta$ for attention/MLP)} & Avg. WER (↓) & Size (↓) \\
    \midrule
    0.99 / 0.999 (b) & 12.6 & 374M \\
    0.99 / 0.998     & 13.3 & 351M \\
    0.99 / 0.997     & 15.8 & 336M \\
    0.99 / 0.996     & 17.3 & 324M \\
    0.99 / 0.995 (c) & 20.1 & 313M \\
    \bottomrule
\end{tabular}
\caption{Further analysis of \whisperT.}
\label{tab:sensitivity-turbo-analysis}
\end{table}

\subsection{Further Analysis of \whisperT}
\label{sec:appndx:turbo-analysis}
In Table \ref{tab:results-acc}, we observe a large jump in WER for \whisperT between configurations (b) and (c). 
Table \ref{tab:sensitivity-turbo-analysis} shows a sensitivity study about finer-grained compression rates between (b) and (c), varying the $\theta$ value for the MLP layers. 
The data indicate that the \whisperT model is sensitive to the degree of compression applied to the MLP layers. The initial step (from 0.999 to 0.998) shows a modest degradation in performance, and further reductions lead to a more significant drop in accuracy.
This trend also suggests that our PCA-based compression method successfully identifies the critical features required to preserve model performance within the given capacity. This demonstrates that the $\theta$ parameter acts as a powerful lever for balancing the trade-off between model compactness and accuracy.

\end{document}